\documentclass{article}
\usepackage{spconf,amsmath,graphicx}
\usepackage{framed}
\usepackage{color}
\usepackage{arydshln}
\usepackage{times}
\usepackage{latexsym}
\usepackage{multirow}
\usepackage{multicol}
\usepackage{graphicx}
\usepackage{booktabs}
\usepackage{subfigure}
\usepackage{tikz}
\usepackage{tikz-qtree}
\usepackage{algorithm}
\usepackage{algorithmic}
\usetikzlibrary{arrows,decorations.pathmorphing,backgrounds,positioning,fit,petri,shapes.misc, arrows.meta,shapes.geometric,decorations.markings,calc,shadows.blur,decorations.pathreplacing,quotes}
\usepackage{pgfplots}
\usepackage{pgfpages}
\usepackage[utf8]{inputenc}
\usepackage{microtype}
\usepackage{inconsolata}
\usepackage{amsmath, amssymb, amsthm}
\usepackage{caption}

\title{ADAPTIVE DATA AUGMENTATION FOR ASPECT SENTIMENT QUAD PREDICTION}
\name{
    Wenyuan Zhang$^{1,2}$, Xinghua Zhang$^{1,2}$, Shiyao Cui$^{*1}$, Kun Huang$^{1,2}$, Xuebin Wang$^{1}$, Tingwen Liu$^{*1,2}$\thanks{*Corresponding author.}
    }
\address{
    $^1$Institute of Information Engineering, Chinese Academy of Sciences, China \\
    $^2$School of Cyber Security, University of Chinese Academy of Sciences, China  
}

\pgfplotsset{compat=1.18}
\begin{document}
\ninept
%
\maketitle
\begin{abstract}
Aspect sentiment quad prediction (ASQP) aims to predict the quad sentiment elements for a given sentence, which is a critical task in the field of aspect-based sentiment analysis.  
However, the data imbalance issue has not received sufficient attention in ASQP task. 
In this paper, we divide the issue into two-folds, quad-pattern imbalance and aspect-category imbalance, and propose  an \textit{\textbf{A}daptive \textbf{D}ata \textbf{A}ugmentation} (ADA) framework to tackle the imbalance issue.
Specifically, a data augmentation process with a condition function adaptively enhances the tail quad patterns and aspect categories, alleviating the data imbalance in ASQP. 
Following previous studies, we also further explore the generative framework for extracting complete quads by introducing the category prior knowledge and  syntax-guided decoding target.
Experimental\footnote{The code is available at  https://github.com/WYRipple/ADA } results demonstrate that data augmentation for imbalance in ASQP task can improve the performance, and the proposed ADA method is superior to naive data oversampling.
\end{abstract}
\section{Introduction}
\label{sec:intro}

\begin{figure}[tb]  
\centering  
\includegraphics
[width=8.5cm]
{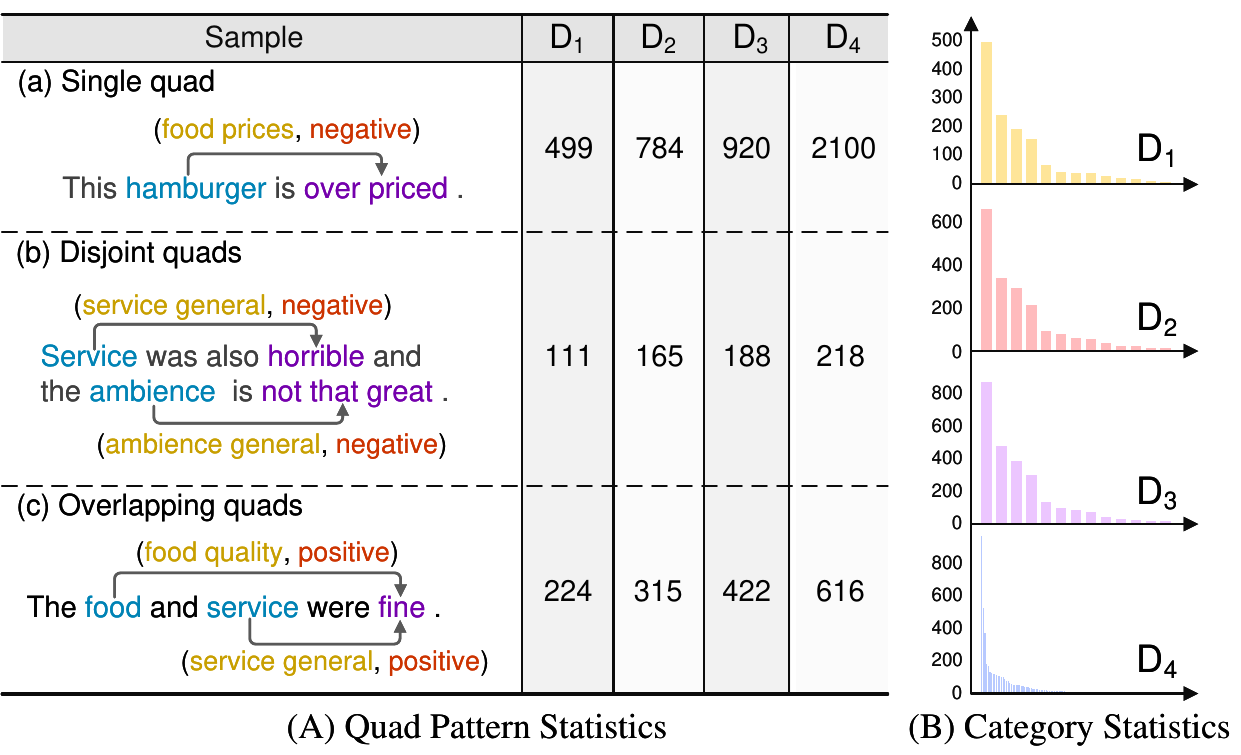}  
\caption{Typical quad patterns with quantity statistics and descending statistical of the number of classes in aspect category, where $\mathtt{D_1}$, $\mathtt{D_2}$, $\mathtt{D_3}$, $\mathtt{D_4}$ respectively denote the public benchmarks~\cite{2021-paraphrase,2021-ACOS} of $\mathtt{Rest15}$, $\mathtt{Rest16}$, $\mathtt{Restaurant}$ and $\mathtt{Laptop}$.}  
\label{motivation}
\end{figure}

Aspect sentiment quad prediction (ASQP)  is  an important task in the field of aspect-based sentiment analysis (ABSA) and has gained increasing research attention \cite{2022-survey,2021-paraphrase,li2023dynamic,2023-exploring}.
The task aims to predict the quad sentiment elements for a given sentence, including {\it aspect term} ($\mathcal{A}$) , {\it opinion term} ($\mathcal{O}$), {\it aspect category} ($\mathcal{C}$)  and {\it sentiment polarity} ($\mathcal{S}$) \cite{2021-paraphrase}.
For example, in (a) of Fig.\ref{motivation} (A), \{``hamburger'' ($\mathcal{A}$), ``over priced'' ($\mathcal{O}$), ``food prices'' ($\mathcal{C}$), ``negative'' ($\mathcal{S}$)\}  constitutes the quad towards the given sentence ``This hamburger is over priced''.

Existing ASQP researches could be roughly grouped into two lines.
The first line of works tag the tokens in the given sentence to derive the quad elements with categories, and are thus considered as \textbf{tagging-based methods}~\cite{2021-ACOS}.
The second line of researches, which are also known as \textbf{generation-based methods} and have become the mainstream of ASQP task, predict the quads by transforming the given sentence into the formalized structure of target quads~\cite{2022-GEN,2022-LEGO,2021-GAS,2021-paraphrase,2023-E2H,2023-mvp,2022-opiniontree}.
For example, Hu et.al~\cite{2022-DLO} formalize the target quads as sequences and improves performances by exploring the order of quad element sequences.
Mao et.al~\cite{2022-seq2path} formalize the quads as tree structures and predict the sentiment elements by generating the paths in the corresponding trees.

Despite of the prior success, the issue of \textbf{data imbalance} are still overlooked.
Specifically, the data imbalance are reflected in two folds, \textbf{quad-pattern imbalance} and \textbf{aspect-category imbalance}.
1) Considering the number and correlations between quads, the patterns in ASQP could be roughly categoried into \textit{single quad}, \textit{disjoint quads} and \textit{overlapping quads} as shown in Fig.\ref{motivation} (A).
Obviously, the quantity to the~\textit{disjoint quads} and~\textit{overlapping quads} patterns are submerged in a large amount of the simple~\textit{single quads}.
With the complexity of these two patterns, the hungry of data  exacerbate the difficulty to their correct predictions.
Hence, such imbalance restricts the overall ASQP performance improvement.
2) The sample quantity to aspect category varies significantly.
As Fig.\ref{motivation} (B) demonstrates, there are a few head aspect categories of adequate samples and the remaining tail aspect categories of limitted samples.
As a result, such data imbalance degrades the ASQP performances across aspect categories.

In this paper, we propose a novel method \textbf{ADA}, which tackles the issues above via \underline{A}daptive \underline{D}ata \underline{A}ugmentation based on a Knowledge-aware Generator.
Specifically, 
we propose a simple yet effective concatenation augmentation method to obtain more challenging augmented samples with diverse themes.
To acquire the appropriate augmentation samples, we propose a condition function which dynamically controls the augmentation process for each quad pattern and aspect category.
Due to the complexity of quad patterns, they are formulated as directed acyclic graphs to guide the augmentation.
We reconstruct the generator as our foundational backbone by introducing prior category knowledge into the input and designing a better syntax-guided target with the incorporation of syntactic knowledge.

Overall, our contributions are as follows:
(1) To the best of our knowledge, we take the lead to explore the data imbalance in ASQP and divide the issue in two-folds: {\it quad-pattern} and {\it aspect-category imbalance}. 
(2) We propose a novel concatenation data augmentation method, which adaptively retrieves samples from the training corpus with a tailor-designed heuristic condition function. In addition, a better generative framework is explored to extract quads in this paper.
(3) Experiments show that our method achieves the state-of-the-art performance on four public benchmarks. Extensive analyses further confirm the effectiveness of our method.

\section{PRELIMINARY}
\subsection{Task Definition}
Given a sentence $X\!\!= <\!\!w_1,\!w_2,\!...,\!w_n\!\!>$ with $n$ words, \textbf{ASQP} aims to extract all quads $\{\mathcal{A},\mathcal{O},\mathcal{C},\mathcal{S}\}$ in \!$X$. $\mathcal{A}$ and $\mathcal{O}$ are aspect and opinion terms respectively which are both textual spans in \!$X$, while aspect category $\mathcal{C}$ and sentiment polarity $\mathcal{S}$ are predefined classes corresponding to $\mathcal{A}$ and $\mathcal{O}$. Each $X$ may have one or multiple quads.

\subsection{Two Types of Imbalanced Phenomena}
\label{sec:two_im}
Data imbalance emerges in many scenarios and has received significant attention~\cite{2020-Decoupling,2017-class-balance}. 
However, it has rarely received attention in ASQP task. 
We first formally summarize two types of data imbalance in this paper: {\it quad-pattern} and {\it aspect-category imbalance}.

\noindent\textbf{Quad-pattern Imbalance.} 
As shown in Fig.\ref{motivation} (A), ASQP pattern can be distributed into three classes: {\it single quad}, {\it disjoint quads} and {\it overlapping quads}.
However, as the multi-terms vary (the quads in a sentence could overlap with each other via the aspect/opinion terms), disjoint or overlapping patterns contain various sub-patterns, which is intractable for implementation.
Hence, we abstract each pattern via a directed acyclic graph (DAG) to fine-grained description and imbalance of quad pattern.
Specifically, let $X^k$ denote the $k$-th sample in training corpus, the DAG for each sample can be formalized as $G^k = \{V^k, E^k\}$:
\begin{equation}
\small
V^k=\{s^k\}\cup\{a^k_{1:i}\}\cup\{o^k_{1:j}\},
\end{equation}
\begin{equation}
\small
E^k=\{s^k\}\times\{a^k_{1:i}\}\cup\{a^k_{1:i}\}\times\{o^k_{1:j}\},
\end{equation}
where $i$ and $j$ are the number of $\mathcal{A}$ and $\mathcal{O}$ in $X^k$. The node set $V^k$ consists of aspect/opinion terms $a^k$/$o^k$, and $s^k$ denotes a virtual node representing sample $X^k$.
The edge set $E^k$ includes the existential dependency relationships from $s^k$-$a^k$ and $a^k$-$o^k$.

\noindent\textbf{Aspect-category Imbalance.} 
Aspect category $\mathcal{C}$ exhibits significant long-tail distributions in all existing ASQP datasets as shown in Fig.\ref{motivation} (B). 
The reason may be that ASQP task tends to analyze the corpus, which originates from customs reviews across various scenarios. (e.g., restaurant or product review), and human customers have review bias. For example, most customers are inclined to review on {\it food quality, price} and {\it service} rather than {\it restaurant location}. Overall, aspect category imbalance is a ubiquitous matter which leads to class bias during training procedure and may impair the performance over all aspect categories.

In implementation, the quad-pattern graph classes are used to depict and retrieve samples for each specific pattern in Section~\ref{sec:ADA}, and each sample $X^k$ has a quad-pattern graph class $p^k$$\gets$$G^k$.
Furthermore, since a single sample may encompass multiple categories, its aspect categories are denoted as $c^k=[c_{1}^k,c_{2}^k,\dots,c_{m}^k]$ and $m$ is the class number of aspect categories in $X^k$.

\begin{figure}[!t]  
\centering  
\includegraphics
[width=8.5cm]
{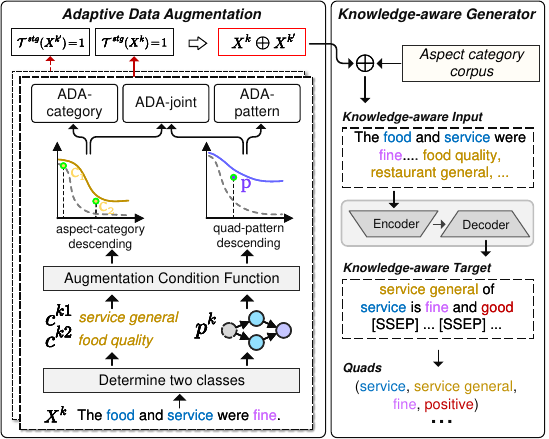}  
\caption{The structure of Knowledge-aware ADA.}
\label{model}
\end{figure}

\section{METHODOLOGY}
Fig.\ref{model} shows our framework which consists of two parts: \textit{\textbf{Adaptive Data Augmentation}} and \textit{\textbf{Knowledge-aware ASQP Generator}}. We describe the tailor-designed data augmentation strategies for two types of data imbalance in Section~\ref{sec:ADA}. To better adapt the generative framework to ASQP task, we develop the Knowledge-aware ASQP generator in Section~\ref{sec:K-a}.

\subsection{Adaptive Data Augmentation}
\label{sec:ADA}

\noindent\textbf{Concatenation Augmentation.} 
Existing text augmentation methods~\cite{2022-MRC,2021-semantics} in ABSA
mainly focus on creating multiple representations of individual sentences,
but ignore the issues of class imbalance mitigation and interrelationships within tuples. 
We propose a augmentation method that generates new samples by concatenating raw samples. 
Its advantages lie in 
(1) 
it can adaptively adjust the proportions of different classes, rather than naively copying (e.g., resampling~\cite{2020-Decoupling});
(2) concatenation operation exhibits a more diverse range of raw sample topics (e.g., food and service), compelling the model to fit the correct semantic relationships within quad elements.
Specifically, let $\mathbb{D}_{raw}$ denote the raw training corpus.
For any samples $X^k$ and $X^{k'}$ in $\mathbb{D}_{raw}$, the concatenation operation will obtain a new sample $X^{kk'}$=$[X^{k}$;$X^{k'}]$=$[X^k_1,...,X^k_{I};X^{k'}_1,...,x^{k'}_{J}]$ with same labels $[y^{k}$;$y^{k'}]$, where $I$ and $J$ are the length of two samples.

\noindent\textbf{Augmentation Condition Function.} 
Based on concatenation augmentation, a condition function is proposed to guide what kind of samples should be augmented.
Taking inspiration from Wang et al.~\cite{2017-class-balance} and Mahajan et al.~\cite{2018-square}, we use negative exponential function to determine the upper bound for augmenting each class, where class sampling probabilities are positively correlated with different class counts.
We calculate the ratio of the current class to the raw maximum class counts, 
when the ratio is less than the upper bound result of the condition function with its raw index as input, this is an augmentable class.
Specifically, the condition functions $\mathcal{F}$ for aspect category and quad-pattern graph of sample $X^k$ are as follows:
\begin{equation}
\small
\mathcal{F}(p^k) = \max(e^{\gamma \cdot pos}+\eta -\frac{n'^p_{pos}}{\kappa n_1^p} ,0),
\label{equ-5}
\end{equation}
\begin{equation}
\small
\mathcal{F}(c^k) = \min(\max(e^{\gamma \cdot pos_{t}}+\eta -\frac{n'^c_{pos_{t}}}{\kappa n_1^c} ,0)|_{t\in[1:m]}),
\label{equ-6}
\end{equation}
where $pos$ represents the index of the current class in the descending array $N^{\varepsilon }$$=$$[n^{\varepsilon }_1,n^{\varepsilon }_2,\dots,n^{\varepsilon }_{N^{\varepsilon}}]$ of $\mathbb{D}_{raw}$, $\varepsilon  \!\in\! \{c,p\}$, where $c$ and $p$ denote the quad-pattern and aspect-category class and $N^{\varepsilon}$ represent the total number of each class, 
$m$ is the same as in Section~\ref{sec:two_im}.
$n_1$ is the raw maximum class count
and $n'$ is the dynamically increasing count of each class after concatenation. 
Parameters $\gamma$ ,$\eta$ and $\kappa$ control the curvature and height of the curve. 
When the class ratio during augmentation does not reach the upper bound,
$\mathcal{F}$ is the condition function with a non-negative scalar output that decides on concatenation augmentation. For aspect category, $min$ will return a positive value only when all aspect categories in a sample meet the requirements.

\noindent\textbf{Augmentation Strategies.} 
Based on the condition function, we propose three augmentation strategies: \textit{\textbf{ADA-pattern strategy}}, \textit{\textbf{ADA-category strategy}}, and \textit{\textbf{ADA-joint strategy}}, which explain how to augment.
We use $\mathcal{T}^{stg}, stg\!\in\!\{P,C,J\}$, to represent above three strategies, which are essentially binary functions based on the result of the condition function feedback, and it can control the concatenation of raw samples.
It should be noted that $\mathcal{T}^J$ in Eq.~\ref{funT} can represent both aspect categories and quad-pattern graphs.
Specifically, for any sample pair $X^k$ and $X^{k'}$ in $\mathbb{D}_{raw}$, the dynamically concatenated dataset based on the three strategies $\mathcal{T}^{stg}$ is denoted as $\mathbb{D}_{con}$:
\begin{equation}
\small
    \mathbb{D}_{con}=\sum_{k=1}^{|\mathbb{D}_{raw}|}\sum_{k'=1}^{|\mathbb{D} _{raw}|} \mathcal{T}^{stg}  (X^k) X^k \oplus \mathcal{T}^{stg}  (X^{k'}) X^{k'},
\label{Daug}
\end{equation}
\begin{equation}
\small
  \begin{cases}
    \text{$   \mathcal{T}^P(X^k)=sgn(\mathcal{F}(p^k))   $},\\
    \text{$   \mathcal{T}^C(X^k)=sgn(\mathcal{F}(c^k))   $},\\
    \text{$   \mathcal{T}^J(X^k)=sgn(\mathcal{F}(p^k)) \! \wedge sgn(\mathcal{F}(c^k))   $},\\
  \end{cases} \\
  \label{funT}
\end{equation}
where $sgn$ means \textit{signum function} and $\wedge$ denotes logical conjunction. 
The text concatenation operation $\oplus$ is only effective when both  $\mathcal{T}^{stg}$=1 in Eq.\ref{Daug}. 
Eq.~\ref{Daug} will iterate to get more possible combinations until no new concatenatable samples exist.

The final augmentation corpus is $\mathbb{D}_{aug}$=$\mathbb{D}_{raw}$+$\mathbb{D}_{con}$. To fully explore the augmentation performance, we implement a naive oversampling~\cite{2020-oversampling} by concatenating samples that all augmented classes have the same count $n_1^{c}$ or $n_1^{p}$.

\subsection{Knowledge-aware Generator}
\label{sec:K-a}

Our backbone is a generative framework which views ASQP as a natural language generation task following previous mainstream methods~\cite{2023-E2H,2021-paraphrase}.
How to design the \textit{input} sequence and \textit{target} output sequence has always been a challenge. 
To build a powerful generative ASQP framework, we develop knowledge-aware input and target output to focus on ASQP task-specific information by introducing prior category knowledge and element constraint within a quad.

\noindent\textbf{Knowledge-aware Input.} 
Inspired by recent prompt learning researches, we propose aggregating descriptions from all aspect categories after sample augmentation.
This introduces category prior knowledge, which helps to provide relevant semantics for out-of-vocabulary (OOV) categories that do not exist in the predefined aspect categories.
Specifically, We construct the final training set $\mathbb{D}_{aug\text{-}c}$ by concatenating the aspect category set $\{c_1, ..., c_{N^c}\}$ as $X_{aug\text{-}c}\! =\! [X_{aug};c_1;...;c_{N^c}]$ for each $X_{aug}$ in $\mathbb{D}_{aug}$.

\noindent\textbf{Knowledge-aware Target.} 
Generative ASQP framework formalizes the target quads \{$\mathcal{A}$, $\mathcal{O}$, $\mathcal{C}$, $\mathcal{S}$\} into text sequence for decoding.
Previous designs for targets include two types: \textit{structured target} and \textit{textual target}. 
Typical \textit{structured targets}, as exemplified in previous studies, include types ``$\mathcal{C}$, $\mathcal{A}$, $\mathcal{O}$, $\mathcal{S}$, \textit{True/False}" \cite{2022-seq2path} or ``[AT] $\mathcal{A}$ [AC] $\mathcal{C}$ [SP] $\mathcal{S}$ [OT] $\mathcal{O}$" \cite{2022-DLO}. 
And \textit{textual target} designs such as ``$\mathcal{C}$ \textit{is} $\mathcal{S}$ \textit{because} $\mathcal{A}$ \textit{is} $\mathcal{O}$" \cite{2021-paraphrase} or ``$\mathcal{C}$ $\mid $ \textit{the} $\mathcal{A}$ \textit{is} $\mathcal{O}$ $\mid $ $\mathcal{S}$" \cite{2022-GEN}.

However, such targets fail to fully utilize the semantic knowledge of natural language or constraint on the content of quads. 
Notice the logical inclusion relationship between $\mathcal{A}$ and $\mathcal{C}$, and the equivalence between $\mathcal{O}$ and $\mathcal{S}$ in semantic expression, We design a logically tight semantic knowledge-aware target: ``$\mathcal{C}$ of $\mathcal{A}$ is $\mathcal{O}$ and $\mathcal{S}$", where replacement words of $\mathcal{C}$ follows \cite{2021-paraphrase}. 
This target allows flexibility in responding to implied emotional expressions, as evidenced by the lack of aspect or opinion term span in a sentence.
When $\mathcal{A}$ is implicit, it is replaced with ``\textit{something}", and if $\mathcal{O}$ is implicit, the ``$\mathcal{O}$ and" for semantic equality is deleted.

\subsection{Training and Inference}
\noindent\textbf{Training Procedure.}
We use $\mathbb{D}_{aug\text{-}c}$ as training set, and employ a pre-trained generative model. We fine-tune the parameters with the input $X_{aug\text{-}c}$ in $\mathbb{D}_{aug\text{-}c}$ and target output $S_{ka}$ in order to maximize the log-likelihood $p_{\theta}(S_{ka}|X_{aug\text{-}c})$, where $\theta$ is initialized with pre-trained weights:
\begin{equation}
\small
\mathcal{L} =\max_{\theta}{\textstyle \sum_{t=1}^{n}}\log p_{\theta}(S_{t,ka}|X_{aug\text{-}c},S_{<t,ka}) ,
\end{equation}
where $n$ is length of sequence $S_{ka}$ and $S_{<t,ka}$ denotes previously generated tokens.

\noindent\textbf{Inference and Quad Recovery.}
We generate $S_{ka}$ by decoding quads in an autoregressive manner, and use a segmentation token "[SSEP]" to distinguish between multiple quads.

\begin{table*}[!t]
\setlength\tabcolsep{6.5pt}
\renewcommand\arraystretch{1}
\small
\footnotesize
    \centering
    \begin{tabular}{l ccc ccc ccc ccc c}
    \midrule
    \multirow{2}{*}{Methods} & \multicolumn{3}{c}{$\mathtt{R15}$} & \multicolumn{3}{c}{$\mathtt{R16}$} & \multicolumn{3}{c}{$\mathtt{Rest}$} & \multicolumn{3}{c}{$\mathtt{Lap}$} &
    \multicolumn{1}{c}{$\mathtt{Avg}$} \\
    & $\mathtt{Pre}$ & $\mathtt{Rec}$ & $\mathtt{F1}$ & $\mathtt{Pre}$ & $\mathtt{Rec}$ & $\mathtt{F1}$ & $\mathtt{Pre}$ & $\mathtt{Rec}$ & $\mathtt{F1}$ & $\mathtt{Pre}$ & $\mathtt{Rec}$ & $\mathtt{F1}$ & $\mathtt{F1}$ \\
    \midrule
    TAS-BERT \cite{2020-TAS}  & 44.24 &28.66 &34.78 &48.65 &39.68 &43.71 & 26.29 &46.29 &33.53 & \textbf{47.15} &19.22 &27.31 & 34.83\\
    Extract-Classify \cite{2021-ACOS}  & 35.64 &37.25 &36.42 &38.40 &50.93 &43.77 & 38.54 &52.96 &44.61 & 45.56 &29.48 &35.80 &40.15\\
    GAS \cite{2021-GAS}  & 45.31 &46.70 &45.98 &54.54 &57.62 &56.04 &57.09 &57.51 &57.30 &43.45 &43.29 &43.37 & 50.67\\
    Paraphrase \cite{2021-paraphrase} &46.16 &47.72 &46.93 &56.63 &59.30 &57.93 &59.85 &59.88 &59.87 &43.44 &42.56 &43.00 &51.93\\
    DLO \cite{2022-DLO} & 47.08 &49.33 &48.18 &57.92 &61.80 &59.79 &60.02 &59.84 &59.93 &43.40 &43.80 &43.60 &52.88\\
    E2H \cite{2023-E2H} & -& -& 49.45 & -& -& 59.55 & -& -& 60.66 & -& -& 43.51 &53.29\\
    DLO+UAUL \cite{2023-uncertainty} & 48.03 &50.54 &49.26 &59.02 &62.05 &60.50 & 61.03 & 60.55 & 60.78 &43.78 &43.53 &43.65 &53.55 \\
    Paraphrase+UAUL \cite{2023-uncertainty} & 48.96 &49.81 &49.38 &58.28 &60.58 &59.40 &60.39 &60.04 &60.21 &44.91 & 44.01 & 44.45 &53.36\\
    One-ASQP \cite{2023-one-step} & -& -& -& -& - & - & \textbf{62.60} &57.21 &59.78 &42.83 &40.00 &41.37 &-\\
    
    \midrule
    oversampling-category &48.69 &51.45& 50.03 & \textbf{59.95} &62.20 & 61.06 &59.93 & 60.13& 60.03 &42.19 &41.17 & 41.67  & 53.20\\
    oversampling-pattern &47.09 & 48.93 & 48.00 & 58.66 & 59.32 &58.99 &56.75 &56.94 & 56.84 &41.39 & 40.40& 40.89 & 51.18\\
    \midrule
    ADA-category & 49.11 &52.20 &50.61 & 59.54 & 64.83 & 62.07 &58.99 &59.51 &59.25 &42.84 &42.29 &42.57 &53.63\\
    ADA-pattern & \textbf{49.43} & \textbf{54.84} &\textbf{52.00} & 59.52 & \textbf{65.33} &\textbf{62.29} &58.67 &59.51 &59.09 &44.14 &41.86 &42.97 &54.09\\
    ADA-joint & 49.31 & 53.96 & 51.53 &59.34 &62.83 &61.03 & 60.15 & \textbf{61.95} &\textbf{61.04} & 45.03 &\textbf{44.53} &\textbf{44.78} & \textbf{54.60}\\
    \midrule
    \end{tabular}
    \caption{Performances of ADA and baselines on four datasets (statistically significant with p $<$ 0.05).}
    \label{table:results}
\end{table*}

\section{EXPERIMENTS}
\subsection{Datasets and Metrics}
We evaluated our method on four publicly available datasets. The SemEval tasks~\cite{2015-semeval,2016-semeval} proposed the $\mathtt{Rest15}$ and $\mathtt{Rest16}$ datasets, which were later completed by \cite{2021-paraphrase}. These two datasets introduce implicit settings for aspect term expressions.
In addition, two publicly available datasets, based on the SemEval 2016 Restaurant dataset \cite{2016-semeval} and annotated data from Amazon,
namely $\mathtt{Restaurant}$ and $\mathtt{Laptop}$ proposed by \cite{2021-ACOS}, place more emphasis on implicit expression of sentiment. Therefore, aspect and opinion terms may all not be explicitly mentioned.
The model performances are evaluated with the metrics of  precision (Pre, \%), recall (Rec, \%) and F1 score (F1, \%), and these datasets can be abbreviated as $\mathtt{\textbf{R15}}$, $\mathtt{\textbf{R16}}$, $\mathtt{\textbf{Rest}}$, and $\mathtt{\textbf{Lap}}$.

\subsection{Comparing Methods and Training settings}
\textbf{Baselines.}
To verify the performance of our method, we compare our results with other state-of-the-art results, \textit{i.e.}, \textbf{TAS-BERT}~\cite{2020-TAS}, \textbf{Extract-Classify}~\cite{2021-ACOS}, \textbf{GAS}~\cite{2021-GAS}, \textbf{Paraphrase}~\cite{2021-paraphrase}, \textbf{DLO}~\cite{2022-DLO}, \textbf{E2H}~\cite{2023-E2H}, \textbf{UAUL}~\cite{2023-uncertainty} and \textbf{One-ASQP}~\cite{2023-one-step}.

\noindent\textbf{Implementation Details.}
We use T5-base\footnote{https://github.com/huggingface/transformers} as the generative PLM, and set \textbf{$\gamma$}, \textbf{$\eta$}, and \textbf{$\kappa$} to [0.05/0.5/2.5, 0.05/0.5/2.5, 0.05/0.4/2, -0.1/0/1] in four datasets.
Our experiments are conducted on a single NVIDIA A100 and based on pytorch-lightning 1.3.5.

\begin{table}[!t]
    \centering
    \small
    \footnotesize
    \setlength\tabcolsep{5pt}
    \renewcommand\arraystretch{1}
    \begin{tabular}{lccc}
    \midrule
    Ablation & &$\mathtt{R15}$ & $\mathtt{R16}$ \\
    \midrule
    \textit{ADA} & \textit{Methods of comparison} & \textbf{51.38} & \textbf{61.80}\\
    \hdashline
    \multirow{2}{*}{K-a Generator} 
    & w/o Data Augmentation & 48.78 & 59.93 \\
    & w/o Data Concatenation & 49.75 & 59.61 \\
    \hdashline
    \multirow{3}{*}{Data Augmentation} 
    & \, w\,\,  K-input  w/o K-target  & 50.30 & 61.31 \\
    & w/o K-input \, w\,\, K-target & 50.30 & 59.80 \\
    & w/o K-input w/o K-target & 49.50 & 60.13 \\
    \midrule
    \end{tabular}
    \captionof{table}{Results of ablation study. `w/o Data Concatenation' represents only replicating samples instead of concatenation, and `w/o K-target' replaces the target based on the initial baseline \cite{2021-paraphrase}.}
    \label{table:ablation}
\end{table}

\subsection{Experimental Results and Analysis}
\textbf{Overall Results.}  
The overall results in Table \ref{table:results} show that ADA consistently outperforms the comparison methods in all four datasets.
While the augmentation through naive oversampling can also achieve a certain level of competitive performance, this approach is challenging to yield better results due to the biased nature of the test corpus.
We observed that ADA-joint strategy achieves better performance in all datasets, but ADA-pattern strategy performs better on $\mathtt{R15}$ and $\mathtt{R16}$. This may be attributed to these two datasets exhibit more pattern bias and a higher disparity in sample class distribution between patterns and categories.

\begin{figure}[!t]
    \centering
    \includegraphics
    [width=7cm]{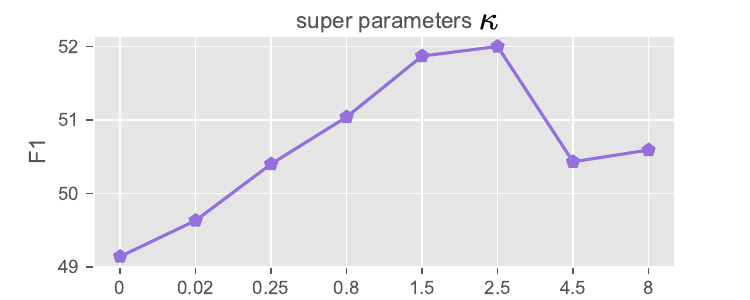}
    \caption{F1 of $\mathtt{R15}$ by setting different parameter $\kappa$.}
    \label{fig:sp}
\end{figure}

\noindent\textbf{Ablation Study.}
The results in Table \ref{table:ablation} are the average F1 scores for the three strategies.
Firstly, the knowledge-aware foundational framework can provide a certain baseline performance, but removing the augmentation components or merely replicating samples significantly compromises performance. 
This demonstrates the effectiveness of our enhancement scheme, especially the concatenation operation.
Furthermore, each component (knowledge-aware input and target) can provide powerful performance improvements.
Ablation studies have shown that the ADA only achieves strongest performances when enhanced by data adaptive augmentation and K-a generator simultaneously.

\noindent\textbf{Hyperparameter  Study.} 
Hyperparameter $\kappa$ determines the strength of data augmentation. $\kappa$ $<$ 1 can be regarded as undersampling consistent with the condition function.
Experiments in Fig.\ref{fig:sp} show that not only the tail, the head augmentation that conforms to the condition function also have a positive impact on performances, but will decline as $\kappa$ increases.

\noindent\textbf{Effect of Imbalanced Classes.} 
Table \ref{table:results2} demonstrates that our approach can significantly enhance the performance of tail classes, particularly in complex patterns or tail aspect categories.
Even though it may slightly affect the performance of head classes (such as \textit{pattern a} in $\mathtt{R16}$), the model gains the ability from more challenging concatenated samples and leads to substantial overall benefits.

\begin{table}[!t]
\setlength\tabcolsep{5pt}
\renewcommand\arraystretch{0.9}
\small
\footnotesize
    \centering
    \begin{tabular}{l ccc ccc}
    \midrule
    \multirow{2}{*}{Options} & \multicolumn{3}{c}{$\mathtt{R15}$} & \multicolumn{3}{c}{$\mathtt{R16}$} \\
    & w/o $\mathtt{ADA}$ & w $\mathtt{ADA}$ & $\Delta$ & w/o $\mathtt{ADA}$ & w $\mathtt{ADA}$ & $\Delta$\\
    \midrule
    cate-head & 51.24 & 53.05 & \textbf{+1.81} & 56.91 & 59.14 & \textbf{+2.23} \\
    cate-tail & 34.08 & 41.01 & \textbf{+6.93} & 36.88 & 40.00 & \textbf{+3.12} \\
    \midrule
    pattern a  & 45.88 & 47.65 & \textbf{+1.77} & 55.92 & 54.68 & -1.24 \\
    pattern b  & 47.73 & 56.35 & \textbf{+8.62} & 63.81 & 73.89 & \textbf{+10.08} \\
    pattern c  & 54.51 & 55.90 & \textbf{+1.39} & 61.37 & 66.80 & \textbf{+5.43} \\
    \midrule
    \midrule
    \multirow{2}{*}{Options} & \multicolumn{3}{c}{$\mathtt{Rest}$} & \multicolumn{3}{c}{$\mathtt{Lap}$} \\
    & w/o $\mathtt{ADA}$ & w $\mathtt{ADA}$ & $\Delta$ & w/o $\mathtt{ADA}$ & w $\mathtt{ADA}$ & $\Delta$\\
    \midrule
    cate-head & 63.32& 66.40& \textbf{+3.08}& 44.94& 47.97& \textbf{+3.03} \\
    cate-tail & 41.23& 44.44& \textbf{+3.21}& 13.04& 14.30& \textbf{+1.26} \\
    \midrule
    pattern a  & 56.85& 57.80& \textbf{+0.95}& 41.60& 42.20& \textbf{+0.60} \\
    pattern b  & 66.27& 67.87& \textbf{+1.60}& 49.73& 52.68& \textbf{+2.95} \\
    pattern c  & 57.99& 61.39& \textbf{+3.40}& 41.56& 46.63& \textbf{+5.07} \\
    \midrule
    \end{tabular}
    \caption{Results of different parts for imbalance. `cate-head/tail' denote the head/tail classes of aspect categories, following the setting with the threshold of 100 in~\cite{2019_CVPR} to distinguish between head and tail.}
    \label{table:results2}
\end{table}

\section{CONCLUSION}
In this work, we study how to improve the performance of ASQP tasks by improving the aspect-category and quad-pattern imbalance. We designe an adaptive data augmentation framework to augment data and propose a Knowledge-aware generator to provide targeted prior knowledge. With the above methods, our model achieves state-of-the-art performances. 
In the future, we plan to integrate Large-Language Models with ASQP task, leveraging the open-world knowledge and logical reasoning abilities to solve the imbalance issue.

\section{AcknowledgeMent}
We would like to thank the anonymous reviewers for their insightful comments and constructive suggestions. 
This research is supported by the National
Key Research and Development Program of China (grant No.2021YFB3100600) and the Youth Innovation Promotion Association of CAS (Grant No. 2021153).

\vfill\pagebreak

\bibliographystyle{IEEEbib}
\bibliography{strings,refs}

\end{document}